\documentclass[11pt,a4paper]{article}
\usepackage[hyperref]{naaclhlt2018}
\usepackage{times}
\usepackage{latexsym}
\usepackage{graphicx}
\usepackage{comment}
\graphicspath{ {images/} }
\usepackage{url}
\usepackage{amssymb}
\usepackage{amsmath}
\usepackage{bm}
\usepackage{multirow}
\usepackage{footmisc}

\aclfinalcopy 

\title{Character-based Neural Networks for Sentence Pair Modeling}

\author{Wuwei Lan \and Wei Xu \\ Department of Computer Science and Engineering \\
Ohio State University\\
  {\tt \{lan.105, xu.1265\}@osu.edu}
}

\date{}

\begin{document}
\maketitle

\begin{abstract}

Sentence pair modeling is critical for many NLP tasks, such as paraphrase identification, semantic textual similarity, and natural language inference. Most state-of-the-art neural models for these tasks rely on pretrained word embedding and compose sentence-level semantics in varied ways; however, few works have attempted to verify whether we really need pretrained embeddings in these tasks. In this paper, we study how effective subword-level (character and character n-gram) representations are in sentence pair modeling. Though it is well-known that subword models are effective in tasks with single sentence input, including language modeling and machine translation, they have not been systematically studied in sentence pair modeling tasks where the semantic and string similarities between texts matter. Our experiments show that subword models without any pretrained word embedding can achieve new state-of-the-art results on two social media datasets and competitive results on news data for paraphrase identification. 

\end{abstract}

\section{Introduction}

Recently, there have been various neural network models proposed for sentence pair modeling tasks, including semantic similarity \cite{agirre-EtAl:2015:SemEval2}, paraphrase identification \cite{dolan2004unsupervised,xu2015semeval}, natural language inference \cite{bowman-EtAl:2015:EMNLP}, etc. Most, if not all, of these state-of-the-art neural models \cite{TACL831,parikh-EtAl:2016:EMNLP2016,he-lin:2016:N16-1,tomar2017neural,shen-yang-deng:2017:EMNLP2017} have achieved the best performances for these tasks by using pretrained word embeddings, but results without pretraining are less frequently reported or noted. In fact, we will show that, even with fixed randomized word vectors, the pairwise word interaction model \cite{he-lin:2016:N16-1} based on contextual word vector similarities can still achieve strong performance by capturing identical words and similar surface context features. Moreover, pretrained word embeddings generally have poor coverage in social media domain where out-of-vocabulary rate often reaches over 20\% \cite{baldwin-EtAl:2013:IJCNLP}. 

We investigated the effectiveness of subword units, such as characters and character n-grams, in place of words for vector representations in sentence pair modeling. Though it is well-known that subword representations are effective to model out-of-vocabulary words in many NLP tasks with a single sentence input, such as machine translation \cite{luong-EtAl:2015:ACL-IJCNLP,costajussa-fonollosa:2016:P16-2}, language modeling \cite{wang:2015,vania-lopez:2017:Long}, and sequence labeling \cite{santos2015boosting,plank-sogaard-goldberg:2016:P16-2}, they are not systematically studied in the tasks that concern pairs of sentences. Unlike in modeling individual sentences, subword representations have impacts not only on the out-of-vocabulary words but also more directly on the relation between two sentences, which is calculated based on vector similarities in many sentence pair modeling approaches (more details in Section \ref{sec:pair_models}). For example, while subwords may capture useful string similarities between a pair of sentences (e.g. spelling or morphological variations: \textit{sister} and \textit{sista}, \textit{teach} and \textit{teaches}), they could introduce errors (e.g. similarly spelled words with completely different meanings: \textit{ware} and \textit{war}). 

To better understand the role of subword embedding in sentence pair modeling, we performed experimental comparisons that vary (1) the type of subword unit, (2) the composition function, and (3) the datasets of different characteristics. We also presented experiments with language modeling as an auxiliary multi-task learning objective, showing consistent improvements. Taken together, subword and language modeling establish new state-of-the-art results in two social media datasets and competitive results in a news dataset for paraphrase identification without using any pretrained word embeddings. 

\begin{table*}[ht!]
\centering
\small
\begin{tabular}{ccccccc}
\hline
Dataset & Training Size& Test Size & \# INV & \# OOV & OOV Ratio & Source \\
\hline
PIT-2015 & 11530 & 838 & 7771 & 1238 & 13.7\% & Twitter trends \\
Twitter-URL & 42200 & 9324 & 24905 & 11440 & 31.5\% & Twitter/news \\
MSRP & 4076 & 1725 & 16226 & 1614 & 9.0\% & news \\
\hline
\end{tabular}
\caption{Statistics of three benchmark datasets for paraphrase identification. The training and testing sizes are in numbers of sentence pairs. The number of unique in-vocabulary (INV) and out-of-vocabulary (OOV) words are calculated based on the publicly
available GloVe embeddings (details in Section \ref{sec:settings}).}
\label{datasets_summary}
\end{table*}

\section{Sentence Pair Modeling with Subwords}

The current neural networks for sentence pair modeling \cite[etc]{TACL831,parikh-EtAl:2016:EMNLP2016,he-lin:2016:N16-1,liu-EtAl:2016:EMNLP20162,tomar2017neural,wang2017bilateral,shen-yang-deng:2017:EMNLP2017} follow a more or less similar design with three main components: (a) contextualized word vectors generated via Bi-LSTM, CNN, or attention, as inputs; (b) soft or hard word alignment and interactions across sentences; (c) and the output classification layer. Different models vary in implementation details, and most importantly, to capture the same essential intuition in the word alignment (also encoded with contextual information) -- the semantic relation between two sentences depends largely on the relations of aligned chunks \cite{agirre-EtAl:2016:SemEval2}. In this paper, we used pairwise word interaction model \cite{he-lin:2016:N16-1} as a representative example and staring point, which reported robust performance across multiple sentence pair modeling tasks and the best results by neural models on social media data \cite{lan2017continuously}.


\subsection{Pairwise Word Interaction (PWI) Model }
\label{sec:pair_models}

Let \(\bm{w}^a= (\bm{w}^a_{1},...,\bm{w}^a_{m})\) and \(\bm{w}^b= (\bm{w}^a_{1},...,\bm{w}^b_{n})\) be the input sentence pair consisting of \( m\) and \(n\) tokens, respectively. Each word vector \(\bm{w}_{i}\) $\in$ \(\mathbb{R}^d\) is initialized with pretrained \(d\)-dimensional word embedding \cite{pennington2014glove, TACL571, wieting2016towards}, then encoded with word context and sequence order through bidirectional LSTMs:
\begin{align}
    &\overrightarrow{\bm{h}}_{i} = LSTM^{f}(\bm{w}_{i}, \overrightarrow{\bm{h}}_{i-1}) \\
    &\overleftarrow{\bm{h}}_{i} = LSTM^{b}(\bm{w}_{i}, \overleftarrow{\bm{h}}_{i+1}) \\
    &\overleftrightarrow{\bm{h}}_{i} = [\overrightarrow{\bm{h}}_{i}, \overleftarrow{\bm{h}}_{i}] \\
    &\bm{h}^{+}_{i} = \overrightarrow{\bm{h}}_{i} + \overleftarrow{\bm{h}}_{i}
\end{align}
\noindent where \(\overrightarrow{\bm{h}}_{i}\) represents forward hidden state, \(\overleftarrow{\bm{h}}_{i}\) represents backword hidden state, and \(\overleftrightarrow{\bm{h}}_{i}\) and \(\bm{h}^{+}_{i}\) are the concatenation and summation of two directional hidden states. 

For all word pairs $(\bm{w}^a_{i}, \bm{w}^b_{j})$ across sentences, the model directly calculates word pair interactions using cosine similarity, Euclidean distance, and dot product over the outputs of the encoding layer:
\begin{align}\label{eq:tensor}
    D(\overrightarrow{\bm{h}}_{i}, \overrightarrow{\bm{h}}_{j}) & = [cos(\overrightarrow{\bm{h}}_{i}, \overrightarrow{\bm{h}}_{j}), \\
    &\qquad L2Euclid(\overrightarrow{\bm{h}}_{i},\overrightarrow{\bm{h}}_{j}), \nonumber\\
    &\qquad DotProduct(\overrightarrow{\bm{h}}_{i},\overrightarrow{\bm{h}}_{j}) ]. \nonumber
\end{align}
\noindent The above equation can also apply to other states \(\overleftarrow{\bm{h}}\), \(\overleftrightarrow{\bm{h}}\) and \(\bm{h}^{+}\), resulting in a tensor \(\mathbf{D}^{13 \times m \times n}\) after padding one extra bias term. A ``hard'' attention is applied to the interaction tensor to further enforce the word alignment, by sorting the interaction values and selecting top ranked word pairs. A 19-layer-deep CNN is followed to aggregate the word interaction features and the softmax layer to predicate classification probabilities. 


\subsection{Embedding Subwords in PWI Model}
Our subword models only involve modification of the input representation layer in the pairwise interaciton model. Let \( {c}_{1},...,{c}_{k}\) be the subword (character unigram, bigram and trigram) sequence of a word $w$. The subword embedding matrix is \( \mathbf{C} \in \mathbb{R}^{d'*k}\), where each subword is encoded into the \(d'\)-dimension vector. The same subwords will share the same embeddings. We considered two different composition functions to assemble subword embeddings into word embedding:

\paragraph{Char C2W} \cite{wang:2015} applies Bi-LSTM to subword sequence \( {c}_{1},...,{c}_{k}\), then the last hidden state \( \overrightarrow{\bm{h}}^{char}_{k}\) in forward direction and the first hidden state \( \overleftarrow{\bm{h}}^{char}_{0}\) of the backward direction are linearly combined into word-level embedding \( \bm{w}\):
\begin{align}
    \bm{w} = \bm{W}_{f} \cdot \overrightarrow{\bm{h}}^{char}_{k} + \bm{W}_{b} \cdot \overleftarrow{\bm{h}}^{char}_{0} + \bm{b}
\end{align}
\noindent where \(\bm{W}_{f}\), \(\bm{W}_{b}\) and \( \bm{b}\) are parameters. 

\paragraph{Char CNN} \cite{kim2016character} applies a convolution operation between subword sequence matrix \( \mathbf{C}\) and a filter \(\mathbf{F} \in \mathbb{R}^{d' \times l}\) of width \(l\) to obtain a feature map \( \mathbf{f} \in \mathbb{R}^{k-l+1}\):
\begin{align}
    \mathbf{f}_j = tanh(\langle\mathbf{C}[\ast, j : j + l-1],\mathbf{F}\rangle + b)
\end{align}
\noindent where \(\langle A, B \rangle = Tr (\bm{A}\bm{B}^{T}) \) is the Frobenius inner product, \(b\) is a bias and \(\mathbf{f}_j\) is the \(j\)th element of \(\mathbf{f}\). We then take the max-over-time operation to select the most important element:
\begin{align}
    y_{\mathbf{f}}=\max_{j} \mathbf{f}_j.
\end{align}
After applying \(q\) filters with varied lengths, we can get the array \(\bm{w}=[{y}_{1},...,{y}_{q}]\), which is followed by a one-layer highway network to generate final word embedding.

\subsection{Auxiliary Language Modeling (LM)}
We adapted a multi-task structure, originally proposed by \cite{DBLP:journals/corr/Rei17} for sequential tagging, to further improve the subword representations in sentence pair modeling. In addition to training the model for sentence pair tasks, we used a secondary language modeling objective that predicts the next word and previous word using softmax over the hidden states of Bi-LSTM as follows:
\begin{align} 
\overrightarrow{E}_{LM} = - \sum^{T-1}_{t=1}(\log(P(w_{t+1}|\overrightarrow{\bm{m}_t})) \\
\overleftarrow{E}_{LM} = - \sum^{T}_{t=2}(\log(P(w_{t-1}|\overleftarrow{\bm{m}_t})) 
\end{align}
\noindent where $\overrightarrow{\bm{m}_t} = tanh(\overrightarrow{\bm{W}}_{hm} \overrightarrow{\bm{h}}_t)$ and $\overleftarrow{\bm{m}_t} = tanh(\overleftarrow{\bm{W}}_{hm} \overleftarrow{\bm{h}}_t)$. The Bi-LSTM here is separate from the one in PWI model. The language modeling objective can be combined into sentence pair modeling through a joint objective function:
\begin{equation} 
E_{joint} = E + \gamma (\overrightarrow{E}_{LM} + \overleftarrow{E}_{LM}),
\end{equation}
\noindent which balances subword-based sentence pair modeling objective $E$ and language modeling with a weighting coefficient \(\gamma\).

\section{Experiments}

\begin{table*}[t]
\footnotesize
\centering
\begin{tabular}{cccccccc}
\hline
 & \multicolumn{2}{c}{Model Variations} & pre-train & \#parameters & Twitter URL & PIT-2015 & MSRP \\
\hline

\multirow{6}{*}{Word Models}  & \multicolumn{2}{c}{Logistic Regression} & -- & -- & 0.683  & 0.645 & 0.829 \\
&  \multicolumn{2}{c}{\cite{lan2017continuously}} & Yes & 9.5M & 0.749  & \underline{0.667} & 0.834 \\
 \cline{2-8}
& \multicolumn{2}{c}{pretrained,  fixed} & Yes & 2.2M & 0.753 & 0.632  & 0.834 \\

& \multicolumn{2}{c}{pretrained, updated} & Yes & 9.5M & 0.756 & 0.656  & 0.832 \\
& \multicolumn{2}{c}{randomized, fixed} & --  & 2.2M & 0.728 & 0.456  & 0.821 \\
& \multicolumn{2}{c}{randomized, updated} & --  & 9.5M & 0.735 & 0.625 & 0.834 \\
\hline
\hline
\multirow{6}{*}{Subword Models} & \multicolumn{2}{c}{C2W, unigram} & --  & 2.6M & 0.742 & 0.534 & 0.816 \\
& \multicolumn{2}{c}{C2W, bigram} & -- & 2.7M & 0.742 & 0.563 & 0.825 \\
& \multicolumn{2}{c}{C2W, trigram} & -- & 3.1M & 0.729 & 0.576 & 0.824 \\
& \multicolumn{2}{c}{CNN, unigram} & -- & 6.5M & 0.756 & 0.589 & 0.820 \\
& \multicolumn{2}{c}{CNN, bigram} & -- & 6.5M & 0.760 & 0.646 & 0.814 \\
& \multicolumn{2}{c}{CNN, trigram} & -- & 6.7M & 0.753 & \underline{0.667} & 0.818 \\
\hline
\hline
\multirow{6}{*}{Subword+LM} & \multicolumn{2}{c}{LM, C2W, unigram} & -- &3.5M & 0.760 & \textbf{0.691} & 0.831 \\
& \multicolumn{2}{c}{LM, C2W, bigram} & -- &3.6M & \textbf{0.768} & 0.651 & 0.830 \\
& \multicolumn{2}{c}{LM, C2W, trigram} & -- &4.0M & \underline{0.765} & 0.659 & 0.831\\
& \multicolumn{2}{c}{LM, CNN, unigram} & -- &7.4M & 0.754 & 0.665 & \textbf{0.840} \\
& \multicolumn{2}{c}{LM, CNN, bigram} & -- & 7.4M& 0.761 & \underline{0.667}& \underline{0.835} \\
& \multicolumn{2}{c}{LM, CNN, trigram}& -- & 7.6M& 0.759 & \underline{0.667} & 0.831 \\
\hline
\end{tabular}
\caption{Results in F1 scores on Twitter-URL, PIT-2015 and MSRP datasets. The best performance figure in each dataset is denoted in \textbf{bold} typeface and the second best is denoted by an \underline{underline}. Without using any pretrained word embeddings, the Subword+LM models achieve better or competitive performance compared to word models.} 
\label{whole_table_results}
\end{table*}


\subsection{Datasets}

We performed experiments on three benchmark datasets for paraphrase identification; each contained pairs of naturally occurring sentences manually labeled as paraphrases and non-paraphrases for binary classification: \textbf{Twitter URL} \cite{lan2017continuously} was collected from tweets sharing the same URL with major news outlets such as @CNN. This dataset keeps a balance between formal and informal language. \textbf{PIT-2015} \cite{Xu-EtAl-2014:TACL,xu2015semeval} comes from the Task 1 of Semeval 2015 and was collected from tweets under the same trending topic, which contains varied topics and language styles. \textbf{MSRP} \cite{dolan2005automatically} was derived from clustered news articles reporting the same event in formal language. Table \ref{datasets_summary} shows vital statistics for all three datasets. 


\subsection{Settings}
\label{sec:settings}

To compare models fairly without implementation variations, we reimplemented all models into a single PyTorch framework.\footnote{The code and data can be obtained from the first and second
author's websites.} We  followed the setups in \cite{he-lin:2016:N16-1} and \cite{lan2017continuously} for the pairwise word interaction model, and used the 200-dimensional GloVe word vectors \cite{pennington2014glove}, trained on 27 billion words from Twitter (vocabulary size of 1.2 milion words) for social media datasets, and 300-dimensional GloVe vectors, trained on 840 billion words (vocabulary size of 2.2 milion words) from Common Crawl for the MSRP dataset. For cases without pretraining, the word/subword vectors were initialized with random samples drawn uniformly from the range [−0.05, 0.05]. We used the same hyperparameters in the C2W \cite{wang:2015} and CNN-based  \cite{kim2016character} compositions for subword models, except that the composed word embeddings were set to 200- or 300- dimensions as the pretrained word embeddings to make experiment results more comparable. For each experiment, we reported results with 20 epochs. 

\subsection{Results}
Table \ref{whole_table_results} shows the experiment results on three datasets. We reported maximum F1 scores of any point on the precision-recall curve \cite{Lipton:2014:OTC:3120306.3120322} following previous work. 

\paragraph{Word Models} The word-level pairwise interaction models, even without pretraining (randomzied) or fine-tuning (fixed), showed strong performance across all three datasets. This reflects the effective design  of the BiLSTM and word interaction layers, as well as the unique character of sentence pair modeling, where n-gram overlapping positively signifies the extent of semantic similarity. As a reference, a logistic regression baseline with simple n-gram  (also in stemmed form) overlapping features can also achieve good performance on PIT-2015 and MSRP datasets. With that being said, pretraining and fine-tuning word vectors are mostly crucial for pushing out the last bit of performance from word-level models.

\begin{table}[bp!]
\footnotesize
\centering
\resizebox{\columnwidth}{!}{\begin{tabular}{|c|c|c|c|c|}
    \hline
    \multirow{2}{*}{Model} & \multicolumn{2}{c|}{INV Words} &  \multicolumn{2}{c|}{OOV Words} \\
    \cline{2-5}
    & \textit{any} & \textit{walking} & \textit{\#airport} & \textit{brexit} \\
    \hline
    \multirow{4}{*}{Word} & \textit{anything} & \textit{walk} & \textit{salomon} & \textit{bollocks}\\
    & \textit{anyone} & \textit{running} & \textit{363} & \textit{misogynistic} \\
    & \textit{other} & \textit{dead} & \textit{\#trumpdchotel} & \textit{patriarchy}\\
    & \textit{there} & \textit{around} & \textit{hillarys} & \textit{sexist} \\
    \hline
    \multirow{4}{*}{Subword} & \textit{analogy} & \textit{waffling} & \textit{@atlairport} & \textit{grexit}\\
     & \textit{nay} & \textit{slagging} & \textit{\#dojreport} & \textit{bret}\\
     & \textit{away} & \textit{scaling} & \textit{\#macbookpro} & \textit{juliet}\\
     & \textit{andy} & \textit{\#hacking} & \textit{\#guangzhou} & \textit{\#brexit}\\
    \hline
    \multirow{4}{*}{Subword} & \textit{any1} & \textit{warming} & \textit{airport} & \textit{\#brexit} \\
    \multirow{4}{*}{+ LM }& \textit{many} & \textit{wagging} & \textit{\#airports} & \textit{brit} \\
    & \textit{ang} & \textit{waging} & \textit{rapport} & \textit{ofbrexit} \\
    & \textit{nanny} & \textit{waiting} & \textit{\#statecapturereport} & \textit{drought-hit}\\
    \hline
    \end{tabular}}
    \caption{Nearest neighbors of word vectors under cosine
similarity in Twitter-URL dataset.}
    \label{tab:nearest_neighbor_words}
\end{table}

\paragraph{Subword Models (+LMs)} Without using any pretrained word embeddings, subword-based pairwise word interaction models can achieve very competitive results on social media datasets compared with the best word-based models (pretrained, fixed). For MSRP with only 9\% of OOV words (Table \ref{datasets_summary}), the subword models do not show advantages. Once the subword models are trained with multi-task language modeling (Subword+LM), the performance on all datasets are further improved, outperforming the best previously reported results by neural models \cite{lan2017continuously}. A qualitative analysis reveals that subwords are crucial for out-of-vocabulary words while language modeling ensures more semantic and syntactic compatibility (Table \ref{tab:nearest_neighbor_words}). 

\begin{table*}[t]
\footnotesize
\centering
\begin{tabular}{cccccccc}
\hline
& Model Variations & CNN$^{19}$ & \#parameters  & \#hours/epoch &  Twitter URL & PIT-2015 & MSRP \\
\hline
\multirow{3}{*}{Word Models} & Logistic Regression & -- & -- & -- & 0.683  & 0.645 & 0.829 \\
\cline{2-8}
& \multirow{2}{*}{pretrained, fixed} &  Yes & 2.2M & 4.5h & 0.753 & 0.632  & \underline{0.834} \\

& &  --  & 1.4M & 3.2h & 0.741 & 0.602  & 0.827 \\
\hline
\hline
\multirow{4}{*}{Subword Models}  & \multirow{2}{*}{C2W, unigram} &  Yes & 2.6M & 5.8h & 0.742 & 0.534 & 0.816 \\
& & --  & 1.4M & 4.6h & 0.741 & 0.655 & 0.808 \\
\cline{2-8}
& \multirow{2}{*}{CNN, unigram} &  Yes & 6.5M & 5.4h & 0.756 & 0.589 & 0.820 \\
&  & --  & 5.3M & 4.2h & \underline{0.759} & 0.659 & 0.809 \\
\hline
\hline
\multirow{4}{*}{Subword+LM}
& \multirow{2}{*}{LM, C2W, unigram} &  Yes & 3.5M & 6.5h & \textbf{0.760} & \textbf{0.691} & 0.831 \\
&  & -- & 2.3M & 5.3h & 0.746 & 0.625 & 0.811 \\
\cline{2-8}
& \multirow{2}{*}{LM, CNN, unigram} &  Yes & 7.4M & 5.8h & 0.754 & \underline{0.665} & \textbf{0.840} \\
& & -- & 6.2M & 4.6h & 0.758 & 0.659 & 0.809 \\
\hline
\end{tabular}
\caption{Comparison of F1 scores between the original PWI model with 19-layer CNN for aggregation and the simplified model without 19-layer CNN on Twitter-URL, PIT-2015 and MSRP datasets. The number of parameters and training time per epoch shown are based on the Twitter URL dataset and a single NVIDIA Pascal P100 GPU.} 
\label{19-layer-cnn_vs_2-layer-mlp}
\end{table*}

\subsection{Combining Word and Subword Representations}
In addition, we experimented with combining the pretrained word embeddings and subword models with various strategies: concatenation, weighted average, adaptive models \cite{miyamoto-cho:2016:EMNLP2016} and attention models \cite{rei-crichton-pyysalo:2016:COLING}. The weighted average outperformed all others but only showed slight improvement over word-based models in social media datasets; other combination strategies could even lower the performance. The best performance was 0.763 F1 in Twitter-URL and 0.671 in PIT-2015 with a weighted average: 0.75 $\times$ word embedding $+$ 0.25 $\times$ subword embedding. 


\section{Model Ablations}
In the original PWI model, He and Lin \shortcite{he-lin:2016:N16-1} performed pattern recognition of complex semantic relationships by applying a 19-layer deep convolutional neural network (CNN) on the word pair interaction tensor (Eq. \ref{eq:tensor}). However, the SemEval task on Interpretable Semantic Textual Similarity \cite{agirre-EtAl:2016:SemEval2} in part demonstrated that the semantic relationship between two sentences depends largely on the relations of aligned words or chunks. Since the interaction tensor in the PWI model already encodes word alignment information in the form of vector similarities, a natural question is whether a 19-layer CNN is necessary.

Table \ref{19-layer-cnn_vs_2-layer-mlp} shows the results of our systems with and without the 19-layer CNN for aggregating the pairwise word interactions before the final softmax layer. While in most cases the 19-layer CNN helps to achieve better or comparable performance, it comes at the expense of $\sim$25\% increase of training time. An exception is the character-based PWI without language model, which performs well on the PIT-2015 dataset without the 19-layer CNN and comparably to logistic regression with string overlap features \cite{eyecioglu-keller:2015:SemEval}. A closer look into the datasets reveals that PIT-2015 has a similar level of unigram overlap as the Twitter URL corpus (Table \ref{tab:character_overlap}),\footref{footnote_1} but lower character bigram overlap (indicative of spelling variations) and lower word bigram overlap (indicative of word reordering) between the pairs of sentences that are labeled as paraphrase.

\begin{table}[bp!]
\footnotesize
\centering
\resizebox{\columnwidth}{!}{\begin{tabular}{l|c|c|c}
    \hline
     & Twitter URL & PIT-2015 & MSRP \\
     \hline
     \#char unigrams in shorter sentence (all) & 67.5 & 36.2 & 109.0 \\
     \#char unigrams in longer sentence (all) & 97.7 & 50.5 & 128.5 \\
     \#char unigrams of the union (all) & 101.1 & 53.0 & 130.0 \\
     \#char unigrams of the intersection  (all)& 64.1 & 33.7 & 107.4 \\
     char unigram overlap (all) & 63.4\% & 63.5\% & 82.6\% \\
     char unigram overlap (paraphrase-only) & 68.8\% & 67.0\% & 84.7\% \\
     \hline
     char bigram overlap (all) & 30.8\% & 33.6\% & 67.4\% \\
     char bigram overlap (paraphrase-only) & 48.2\% & 42.4\% & 71.6\% \\
     \hline
     word unigram overlap (all) & 13.3\% & 21.7\% & 54.8\% \\
     word unigram overlap (paraphrase-only) & 32.0\% & 30.2\% & 59.1\% \\

     \hline
     word bigram overlap (all) & 5.3\% & 8.4\% & 33.2\% \\
     word bigram overlap (paraphrase-only) & 17.9\% & 12.3\% & 36.8\% \\
     \hline
    \end{tabular}}
    \caption{Character and word overlap comparison.}
    \label{tab:character_overlap}
\end{table}


The 19-layer CNN appears to be crucial for the MSRP dataset, which has the smallest training size and is skewed toward very high word overlap.\footnote{See more discussions in \cite{lan2017continuously}. \label{footnote_1}} For the two social media datasets, our subword models have improved performance compared to pretrained word models regardless of having or not having the 19-layer CNN.

\section{Conclusion}

We presented a focused study on the effectiveness of subword models in sentence pair modeling and showed competitive results without using pretrained word embeddings. We also showed that subword models can benefit from multi-task learning with simple language modeling, and established new start-of-the-art results for paraphrase identification on two Twitter datasets, where out-of-vocabulary words and spelling variations are profound. The results shed light on future work on language-independent paraphrase identification and multilingual paraphrase acquisition where pretrained word embeddings on large corpora are not readily available in many languages.

\section*{Acknowledgments}
We thank John Wieting for valuable discussions, and Ohio Supercomputer Center \cite{Oakley2012} for computing resources. Funding was provided by the U.S. Army Research Office (ARO) and Defense Advanced Research Projects Agency (DARPA) under Contract Number W911NF-17-C-0095. The content of the information in this document does not necessarily reflect the position or the policy of the Government, and no official endorsement should be inferred. The U.S. Government is authorized to reproduce and distribute reprints for government purposes notwithstanding any copyright notation here on.


\bibliography{naaclhlt2018}
\bibliographystyle{acl_natbib}

\end{document}